# FACE RECOGNITION USING MAP DISCRIMINANT ON YCBCR COLOR SPACE

**I Gede Pasek Suta Wijaya**

Electrical Engineering Department, Engineering Faculty, Mataram University.
Jl. Majapahit 62 Mataram, West Nusa Tenggara, Indonesia.
*Email: gpsutawijaya@te.ftunram.ac.id*

**Abstrak**

This paper presents face recognition using maximum a posteriori (MAP) discriminant on YCbCr color space. The YCbCr color space is considered in order to cover the skin information of face image on the recognition process. The proposed method is employed to improve the recognition rate and equal error rate (EER) of the gray scale based face recognition. In this case, the face features vector consisting of small part of dominant frequency elements which is extracted by non-blocking DCT is implemented as dimensional reduction of the raw face images. The matching process between the query face features and the trained face features is performed using maximum a posteriori (MAP) discriminant. From the experimental results on data from four face databases containing 2268 images with 196 classes show that the face recognition YCbCr color space provide better recognition rate and lesser EER than those of gray scale based face recognition which improve the first rank of grayscale based method result by about 4%. However, it requires three times more computation time than that of grayscale based method.

## 1. Introduction

Human face recognition is an active research area in image processing applications because there are many potential applications, which cover human computer interactions, forensics, surveillance, and security systems. The published approaches mostly related to our system are frequency analysis-based face recognitions, which are combined with PCA and LDA as described in Refs. [1-4]. However, they have to retrain all face image classes to get optimal projection when new classes are added into the system. Moreover, those systems work in grayscale domain.

This research proposes a face recognition using maximum a posteriori (MAP) discriminant on YCbCr color space. The YCbCr color space is considered in order to cover the skin information of face image on the recognition process. In addition, it is employed to improve the recognition rate and equal error rate (EER) of the gray scale based face recognition[5]. The chrominance components (Cb and Cr) have to be considered in the face recognition because the face image mostly covers with skin and skin-tone color which can be modeled by the chrominance elements.

Moreover, the chrominance components of skin-tone color are nonlinearly dependent on the luminance [6].

This paper is organized as follows: section 2 describes the most related work to the proposed method; section 3 explains the algorithm of classical LDA-based face recognition and its weakness; section 4 explains the MAP discriminant and its advantages; section 5 explains how to implement the frequency analysis in the face image in order to obtain the face features; section 6 describes the recognition algorithm; section 7 presents the experimental results and compares our results to LDA and eigen-face method and also describes the results discussion, and the rest concludes the paper.

## 2. Related Work

The most related approach to our system is face recognition based on LDA and its variations as described in Ref. [1-5]. Ref. [1] presented PCA and LDA algorithm in DCT domain. The DCT was implemented to reduce data dimensional of face image[2]. In addition, the Ref. [2] implemented the combination of DCT analysis and face localization techniques for finding the global information of face image, but it requires eyes coordinate which have to input manually to perform geometrical normalization.

Ref. [3] implemented the wavelet transforms to reduce the dimension of face image, employed a regulation scheme for the within-scatter matrix, and used optimization procedure. It was reported that the Daubechies (Db6) was implemented to filter image to resolution 29 x 23. Ref. [4] proposed Direct LDA (DLDA) to overcome the singularity problem LDA. However, those approaches still need high memory space and must be retrained when new face class is added. The Ref. [5] proposed the modification of LDA for solving the retraining problem of LDA, but it just worked on the grayscale face images.

In this method, we consider the chrominance component of face images on recognition process and also implement the frequency analysis (i.e. DCT) for reducing the original data dimensional. In addition, the MAP discriminant is employed for classifying the face class without localization and bounding box processing.

## 3. LDA for Face Recognition

The main purpose of LDA analysis is to find a linear transformation such that feature clusters are most separable after the transformation. It can be achieved by the between-class scatter matrix $S_b$ and the within-class scatter matrix $S_w$ analysis, as explained in Ref. [1,5]. The class separation is measured by the ratio of determinant of the $S_b$ matrix to the $S_w$ matrix using the equation below.

$$E = \arg\max_{E} \frac{|E^T S_b E|}{|E^T S_w E|} \quad (1)$$

Where $E = \{e_1, e_2, e_3, ..., e_m\}$ is set of eigen-vectors corresponding to $m$ largest eigen-values $\lambda_i$ which satisfy the equation: $S_b e_i = \lambda_i S_w e_i$, $i = 1, 2, 3, ..., m$. The eigen-vectors and eigen-values are obtained by computing the inverse of $S_b$ and then solving the eigen problem of $S_W^{-1} S_b$ matrix. Finally, the projection of the linear discriminant functions is given by:

$$Y_i(C) = E^T (C_i - \overline{m_i}) \quad (2)$$

The intrinsic problem of the LDA algorithm is the singularity problem of scatter matrix due to the high data dimensional and small number of training samples. Some methods have been proposed to solve that problem as described in Ref. [1,4]. However, those methods can only reduce the large computational load and memory space requirement, but the retraining problem is not covered yet.

## 4. MAP Discriminant Analysis

In order to avoid retraining problem, the MAP is considered which is based on assumption that the matrix scatter has small dimension and the covariance of the training images is multivariate normal distribution.

Suppose we have a set of face training containing $n$ classes with each class consisting $m$ column vectors of face features. From this set, mean of each class is easily determined and then placed it into a mean matrix as $M = \{\mu_1, \mu_2, \mu_3, ..., \mu_c\}$. Finally, we can determine the global covariance using the with-in class ($S_w$) equation.

$$S_w = C_g = \sum_{i=1}^{c} \sum_{x_j \in C_i} (x_j - \mu_i)(x_j - \mu_i)^T \quad (3)$$

If $C_g$ is multivariate normal distribution, we can classify each face feature to person's class using the equation below.

$$F_c = max[g_1(x), g_2(x), g_3(x), ..., g_m(x)] \quad (4)$$

Where $g_i(x)$ is given by:

$$g_i(x) = \mu_i C^{-1} x^T - 0.5 \mu_i C^{-1} \mu_i^T \quad (5)$$

The equation (5) is derived from maximum a posteriori (MAP) discriminant, as described below:

$$g_i(x) = P(\omega_i / x) = \frac{P(x/\omega_i) P(\omega_i)}{P(x)} \quad (6)$$

$$g_i(x) = \frac{1}{(2\pi^{n/2})|C_i|^{1/2}} \exp\left(-1/2(x-\mu_i)^T C_i^{-1}(x-\mu_i)\right) P(\omega_i) \frac{1}{P(x)} \quad (7)$$

where $P(x)$ is total probability of x. By assuming all classes have the same covariance and prior probability, the equation (5) can be simplified as below:

$$g_i(x) = -\tfrac{1}{2} x C^{-1} x^T + \mu_i C^{-1} x^T - \tfrac{1}{2} \mu_i C^{-1} \mu_i^T \quad (8)$$

By keeping just the terms dependent on $\mu_i$ and $C$, the equation (5) is obtained.

This algorithm has some advantages for classifying the face images classes:
1. It is simple because it does not require the eigen-values and eigen-vectors.
2. It can solve the retraining problem. It can be illustrated: firstly, when a new class added into the system, the MAP just calculates the mean and the covariance of its class; secondly, the newest mean is placed into the matrix M; and finally, the previous covariance is updated by adding it with the newest class covariance.
3. The computation complexity is much less than that of the PCA and LDA because of not requiring eigen analysis.

The weakness of this classification algorithm is singularity problem due to high data dimensional and small number of training samples. To overcome it, we implement frequency analysis to decrease the original data dimensional as explained in the next section.

In order to deal with the color information of face image, those processes are accessed three times which correspond to YCbCr color space then the decision rule is performed by maximum-mean distance which is a modification of equation (4).

$$F_c^{Clr} = \max\{\bar{g}_1(x), \bar{g}_2(x), \bar{g}_3(x), ..., \bar{g}_m(x)\} \quad (9)$$

where $\bar{g}_i(x)$ is mean of ($g_i(x_Y)$, $g_i(x_{Cb})$, $g_i(x_{Cr})$).

## 5. Face Features Extraction

In this research, we employ a holistic approach for face feature extraction based on frequency analysis of an entire face. Our approach is different from the most closely related approach presented in Ref. [1,2]. Ref. [1] implemented 2D-DCT for blocked image as performed in JPEG compression. Ref [2] implemented 2D-DCT for an entire normalized face and kept a small number of the DCT coefficients, having large magnitudes to represents face features. Ref [3] implemented wavelet sub band to create face features. However, our approach implements a 1-dimensional DCT analysis in the entire image without geometrical normalization and bounding process.

In order to generate frequency-matrix of the face image in DCT domain, the 1D-DCT algorithms is implemented twice: on the rows and the columns of the face image matrix. The pseudo code of DCT decomposition of the face image is shown below:

```
func dctDecompose(I:array[0 … r-1, 0 .. c-1])
  for row = 1 to r do
    F[row, 0 .. c-1]=1D-DCT(I[row, 0 .. c-1])
  end for
  for col = 1 to c do
    F[0 .. r-1,col]=1D-DCT(F[0 .. r-1,col])
  end for
  return F
end func
```

Where *I* is face image matrix and *F* is DCT domain of frequency matrix. From the F, the compact and meaningful face features describing holistic information of face image is created by three steps: first, convert the frequency domain matrix to a vector using row ordering technique; second, sort the vector descending order using quick sort algorithm, finally select a small number of vector elements which have large magnitude value (i.e., less then 100 elements).

## 6. The Proposed algorithm

There are three main processes in the proposed method: face features extraction, training, and recognition processes. The face features extraction unit consists of color space transformation, equalization, and frequency analysis. Actually, there are two main functions of it: firstly, to decrease the effect of non-uniform lighting condition on the face image, which is performed using standard equalization, secondly, to create compact face features using frequency analysis (i.e. using DCT analysis). In this case, we do not apply blocked DCT as performed in JPEG compression but non-blocked DCT. In addition, wavelet analysis also can be used to extract the face features. The wavelet analysis usually performs the face analysis using two bases function (DB4). However, the DCT based features extraction is employed for creating the face features, because of much energy compactness.

Those processes are performed on both training and query (probe) face images. However, in the training process, those are performed one time. Finally, face classification is done using MAP discriminant.

## 7. Experiment and Results

The experiments are carried out using data from five face databases: ITS-Lab. Kumamoto University database, INDIA database, EE-UNRAM database, ORL face database, and cropped face databases.

The ITS-Lab database consists of 48 people and each person has 10 pose orientations as shown in Fig. 2. The face images were taken by Konica Minolta camera series VIVID 900 under varying lighting condition. The EE-UNRAM database consists of 40 people and each person has 8 pose orientations: looking front, looking left about $30^0$, looking right about $30^0$, looking up, looking down, and wearing accessory such as glasses. The Indian database consists 61 people (22 women and 39 men), each person has eleven pose orientations: looking front, looking left, looking right, looking up, looking up towards left, looking up towards right, and looking down. Indian database also included the emotions: neutral, smile, laughter, sad/disgust. The ORL database was taken at different times, under varying the lighting conditions, facial with different expressions (open/closed eyes, smiling/not smiling) and facial details (glasses/no glasses). All of the images were taken against a dark homogeneous background. The faces of the subjects are in an upright, frontal position (with tolerance for some side movement). The ORL database is a grayscale face database that consists of 40 people, mainly male. The cropped face databases is open database, consisted of 15 face pose variations, some of them also included pose accessories such as glasses. The example of face pose of each database is shown in Fig. 1.

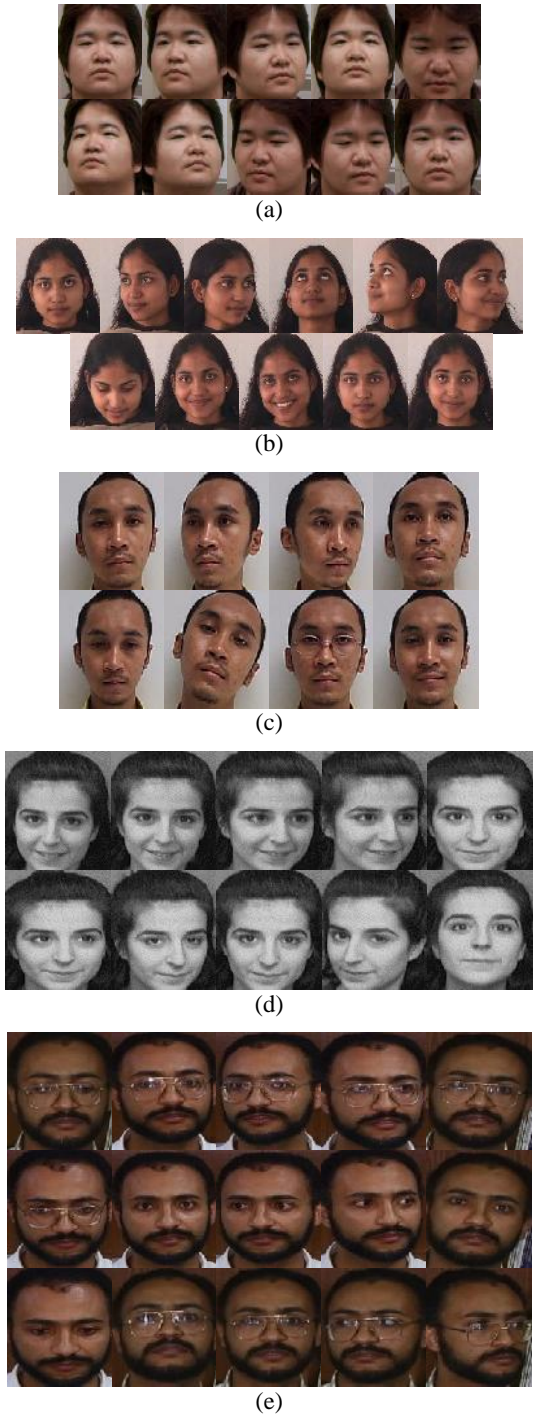

**Fig. 1.** Example of face poses: (a) ITS-Lab., (b) INDIA, (c) UNRAM, (d) ORL and (e) Cropped face databases[5,7-9].

The first experiments was carried out on four face databases: ITS-Lab, INDIA, EE-UNRAM, and ORL. The main aim of these tests were to know the performance of MAP based face recognition method compared to PCA and LDA based face recognition.

In addition, the experimental results shows the MAP based face recognition provide better recognition rate than that of PCA and LDA (see Fig. 2). It can be achieved because the MAP based face recognition performance the classification using the mahalanobis distance while that of PCA and LDA using Euclidean distance.

The second experiment was performed to investigate the effect of considering the chrominance component (using YCbCr color space) of face image on MAP based face recognition. In this test, the ORL face database was not included because it is grayscale database. The experimental results show that the features with considering the chrominance components of face image could provide much higher recognition rate than that of without chrominance (grayscale) as shown in Fig. 3. This achievement means that the chrominance of face image also contains much discriminant information, which cannot be discarded from the recognition process.

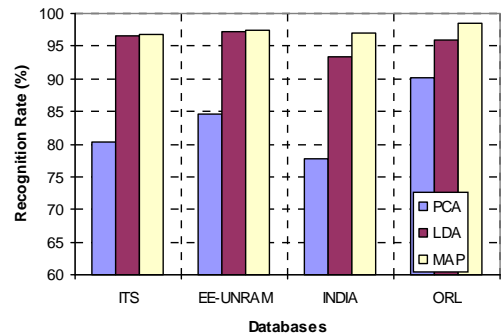

**Fig. 2.** The recognition rate of the MAP based face recognition method compared with the most related methods.

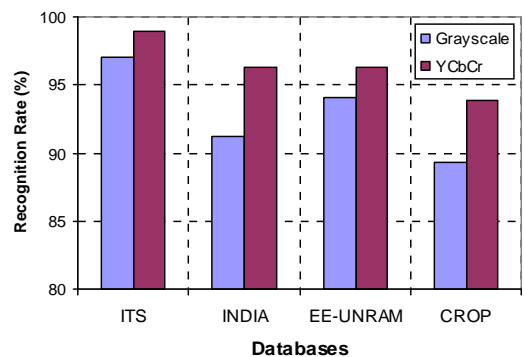

**Fig. 3.** The recognition rate comparison of the proposed method on both grayscale and YCbCr color space.

The third experiment is performed to investigate deeply the effect of chrominance components on recognition rate. The experiments were done on data from ITS, INDIA, UNRAM, and Cropped databases consisting of 2268 face images of 196 classes and the performance evaluation using cumulative matching score (CMS). The result shows that the chrominance components also give higher recognition rate than that of without chrominance components. In term of rank one, the chrominance components provide about 4% higher than that of our previous method (grayscale), as shown in Fig. 4. It means this achievement inline with the previous experimental result, which also prove that the chrominance component of face image cannot be discarded from the recognition process.

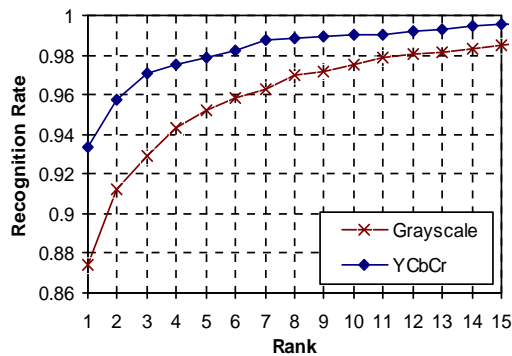

**Fig. 4.** The CMS performance of the proposed method on combine databases.

The last experiments were done to know the accuracy of the proposed method compared to grayscale based (base line) performance. The result shows that chrominance components also provide higher accuracy than that of the base line (see Fig. 5), which is indicated by lower EER (about 0.0457). It also supports the previous achievement that the chrominance components provide much useful information for face recognition.

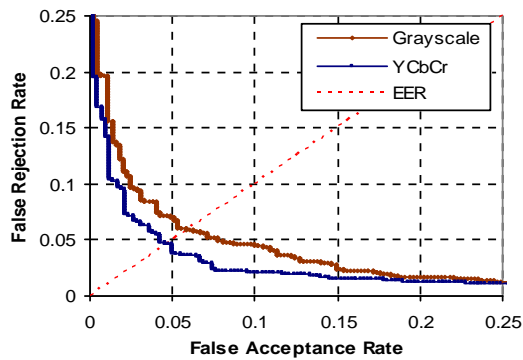

**Fig. 5.** ROC of chrominance components on combined database.

Regarding to time consumption the proposed system requires longer training and querying times (almost three times of the baseline) because it has to access three time of the baseline recognition system for processing three-color elements of face image. However, this system can solve retraining problem of the LDA and PCA based face recognition, as proved experimentally in the Ref. [5].

The chrominance information tends to provide better result than that of grayscale based recognition because the skin color information of face image is included in the recognition process. It means that the skin color is important information of face image, which exist in the chrominance elements. Moreover, human being can visually distinguish face image class using just skin color.

## 8. Conclusions

The MAP based face recognition could provide better recognition rate than that of PCA and LDA based face recognition without requiring the retraining all data samples when new data is added into the system. The chrominance component (CbCr color space) of face image containing much useful discriminant information, which have to be considered on recognition process. It could improve significantly the recognition rate of that of grayscale by about 4% in rank one with the small enough EER (0.0457). However, the color information required longer processing time than that of the grayscale.

In future, the proposed recognition system will be tested in large face database (FERET) to know further effect of chrominance components on recognition rate.

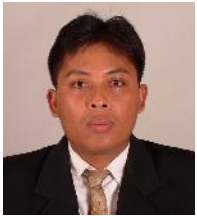
**I Gede Pasek Suta Wijaya** received the B.S. degree in Electrical Engineering from Gadjah Mada University in 1997, M.Eng. degree in Computer Informatics System from Gadjah Mada University in 2001, and D.Eng from Department of Electrical Engineering and Computer Science, Kumamoto University, Japan. During 1998-1999 he worked in Toyota Astra Motor Company in Indonesia as Planning Production Control, and from 1999-2000, next, he worked as lecturer assistance in Yogyakarta National Technology College in Indonesia, and since 2000 he has been full time lecturer and has stayed in Informatics Systems Laboratory in Electrical Department, Mataram University Indonesia. Since 2001-2006 he got research grant form National Education ministry on image processing application